\documentclass[runningheads]{llncs}
\usepackage{graphicx}
\usepackage{tikz,pgfplots}
\pgfplotsset{compat=1.11}
\usepackage{comment}
\usepackage{amsmath,amssymb} % define this before the line numbering.
\usepackage{color}
\usepackage{multirow}
\usepackage{hyperref}
\usepackage[utf8]{inputenc} % arxiv admin
\hypersetup{
    colorlinks=true,
    linkcolor=blue,
    filecolor=cyan,      
    urlcolor=magenta,
}

\begin{document}
% \renewcommand\thelinenumber{\color[rgb]{0.2,0.5,0.8}\normalfont\sffamily\scriptsize\arabic{linenumber}\color[rgb]{0,0,0}}
% \renewcommand\makeLineNumber {\hss\thelinenumber\ \hspace{6mm} \rlap{\hskip\textwidth\ \hspace{6.5mm}\thelinenumber}}
% \linenumbers
\pagestyle{headings}
\mainmatter
\def\ECCVSubNumber{22}  % Insert your submission number here

\title{Curriculum Learning for\\Recurrent Video Object Segmentation} % Replace with your title

% INITIAL SUBMISSION 
%\begin{comment}
\titlerunning{ECCV-20 submission ID \ECCVSubNumber} 
\authorrunning{ECCV-20 submission ID \ECCVSubNumber} 
\author{Anonymous ECCV submission}
\institute{Paper ID \ECCVSubNumber}
%\end{comment}
%******************

% CAMERA READY SUBMISSION
%\begin{comment}
\titlerunning{Curriculum Learning for Recurrent Video Object Segmentation}
% If the paper title is too long for the running head, you can set
% an abbreviated paper title here
%
\author{Maria Gonzàlez-i-Calabuig\inst{1}%\orcidID{0000-1111-2222-3333} 
\and
Carles Ventura\inst{2}%\orcidID{1111-2222-3333-4444} 
\and
Xavier Giró-i-Nieto\inst{1}%\orcidID{2222--3333-4444-5555}
}
\authorrunning{M. Gonzalez-i-Calabuig et al.}
% First names are abbreviated in the running head.
% If there are more than two authors, 'et al.' is used.
%
\institute{Universitat Politècnica de Catalunya
\email{\{maria.gonzalez.calabuig,xavier.giro\}@upc.edu} \and
Universitat Oberta de Catalunya
\email{cventuraroy@uoc.edu}}
%\end{comment}
%******************
\maketitle

\begin{abstract}
Video object segmentation can be understood as a sequence-to-sequence task that can benefit from the curriculum learning strategies for better and faster training of deep neural networks.
This work explores different schedule sampling and frame skipping variations to significantly improve the performance of a recurrent architecture.
Our results on the car class of the KITTI-MOTS challenge indicate that, surprisingly, an inverse schedule sampling is a better option than a classic forward one. Also, that a progressive skipping of frames during training is beneficial, but only when training with the ground truth masks instead of the predicted ones.
Source code and trained models are available at \url{http://imatge-upc.github.io/rvos-mots/}. 
%\footnote{\url{http://imatge-upc.github.io/rvos-mots/}}

%Recent work on the field of video object labeling has shown interesting results when training with multigrid strategies. Our work focuses on two aspects of these multigrid strategies: the impact of a curriculum learning strategy and the performance of the model when using different sizes of image resolution, batch size and clip length; all applied to video object segmentation, a closely connected task to video object labeling. Our work presents the results obtained when implementing two curriculum learning strategies using a recurrent neural network in the KITTI-MOTS benchmark: schedule sampling and frame skipping. 

\end{abstract}

\section{Introduction}

The optimization process of deep neural networks is greatly influenced by how training data is used.
Curriculum learning~\cite{curriculum_learning} is a training strategy for machine learning that consists of presenting simple concepts to the model first to, gradually, increasing their complexity. 
%Introduced in 2009 by Bengio et al. \cite{curriculum_learning}, curriculum learning describes a methodology of training where the examples are not randomly presented to the model, instead, they are organized in a meaningful way. Simple concepts are first presented and gradually become more complex.¡

Our work proposes two training curriculums for a Recurrent Video Object Segmentation engine (RVOS) ~\cite{rvos}, a neural model for one-shot (or semi-supervised) video object segmentation (VOS).
In this task, a binary mask of an object is provided for a single frame and the goal is predicting the mask of the selected object across the rest of the frames in the video sequence. 
RVOS architecture is based on an end-to-end recurrent Conv-LSTM~\cite{xingjian2015convolutional} decoder that tracks objects across frames, with no need of any post-processing. 
The recurrent architecture makes RVOS a fast solution for the task, capable of processing more than 20 frames per second~\cite{athar2020stem}. 
RVOS was originally tested on the DAVIS and YouTube-VOS datasets for one-shot video object segmentation.
We show how RVOS struggles with the \textit{cars} in the KITTI-MOTS dataset~\cite{voigtlaender2019mots}, whose videos are more crowded and challenging than DAVIS or YouTube-VOS.
We improve the off-the-shelf RVOS baseline by modifying its training curriculum in two ways.
First, with a schedule sampling~\cite{bengio2015scheduled} totally contrary to the one original one in RVOS and, secondly, by gradually increasing the complexity of the task by subsampling video frames at training time. 

%The developed source code and trained models are available on our project website \url{http://imatge-upc.github.io/rvos-mots/}. 
%\footnote{\url{http://imatge-upc.github.io/rvos-mots/}}

%Video object segmentation is still a very challenging task in the research community. During the past few years, this task has been gaining more attention due to the introduction of deep learning algorithms and the apparition of new and more complete datasets. Also, the increment in video data due to the rapid development of the Internet and mobile telephones has put video object segmentation on the spotlight in order to be able to analyze all this data efficiently. Among the different architectures and techniques that address this task, the concept of curriculum learning appears. 

%On the literature, a wide range of strategies that apply the concept behind this technique can be found. This paper makes focus on two of them: schedule sampling and frame skipping. These techniques are implemented on a recurrent neural network model, RVOS \cite{rvos}, using the KITTI-MOTS dataset \cite{KITTI}. It is demonstrated that curriculum learning strategies increase the model performance on recurrent neural networks.  

\section{Related Work}

\textbf{Schedule Sampling}~\cite{bengio2015scheduled} offers an alternative to \textit{teacher forcing}~\cite{teacher_forcing} where, during training time, the model has access to the ground truth label of the previous time-step in each new prediction.
During inference, the model uses its predictions as input in the next training step. This may lead to exposure bias because of the discrepancy between training and inference and result in poor model performance.
Schedule sampling takes benefit from teacher forcing while avoiding exposure bias by gradually replacing the ground-truth tokens by the model’s predictions. 
%The model is forced to learn to deal with its own mistakes as it would during inference.
Three different decay schedules were proposed by Bengio et al.~\cite{bengio2015scheduled}: exponential, inverse sigmoid and linear.
While Ren and Zemel~\cite{ren2017end} and Xu et al.~\cite{xu2018youtube} used a linear schedule in their training, Oh et al.~\cite{wug2018fast} and RVOS~\cite{rvos} adopted a more drastic scheme, using ground truth labels in the first half of the training, and predicted masks in the second half. We have named this second approach as a \textit{step} schedule, as in the well-known Heaviside step function.

\textbf{Frame Skipping} is a training curriculum in which video sequences are progressively sub-sampled in time so that the model is exposed to sequences with faster changes, even if synthetically generated. 
%This technique is partially motivated by the tight constraints in terms of memory resources when training deep neural networks with video sequences.
The limited sizes of the mini-batches typically force training with short sequences which, in the case of video, may be highly redundant if considering consecutive frames.

Frame Skipping was introduced in the Space-Time Memory Networks (STM)~\cite{oh2019video}, inspired by 
%a previous work on 3D reconstruction with RNNs
~\cite{yang2015weakly} and related to their own previous model\cite{oh2019fast}. 
%trained by randomly removing frames in the training sequences. 
STMs increase gradually the amount of skipped frames, from 0 to 25. %Their attention-based architecture is trained with clips of only length 3.
Wu et al.~\cite{Wu_2020_CVPR} achieved relevant gains when processing video streams at a \textit{fast} and a \textit{slow} frame rates in two different pathways that merge at the deepest layer. 

%While we observe a gain in training at different frame rates, we keep a single pathway and introduce the

%are grounded in this concept, by combining training batches with different spatial-temporal resolutions that changed according to a schedule. 

%limitations of the number of consecutive frames that the model can see on one iteration due to memory constraints. 
%This data augmentation 
%which are to "see" more changes on slow video sequences and to make the model more robust to these changes.   
%choosing non-consecutive frames of a video sequence, skipping a specific number of frames to allow the model to "see" more changes on slow video sequences and to make the model more robust to these changes. 
%addresses the high redundancy and low entropy between consecutive frames in a video sequencet 
%This strategy appears motivated 
%This strategy 

%The state of the art for supervised video object segmentation include a frame skipping approach in their works~\cite{foh2019video,oh2019fast,frame_skipping_3}, rely on this strategy on their different works that address video object segmentation. They randomly skip frames during sampling in order to learn the appearance change over a long time. 

%On \cite{frame_skipping_different}, another example is presented where the authors present three implementations of this technique.

%, where they present a multigrid strategy, for each technique, two sets of experiments are presented to study the impact of different configurations when training. On the first set, smaller and compressed images with a resolution of 256x448 are used for training. 

\section{Experiments}

We have explored different schedule sampling and frame skipping strategies with the RVOS model~\cite{rvos} evaluated on the \textit{car} class in the validation partition of the KITTI-MOTS benchmark~\cite{voigtlaender2019mots}. The task addressed is the one-shot (or semi-supervised) video object segmentation (VOS) task, where a mask of the object
%the first ground-truth annotation of the frame in which the instance appears for the first time 
is provided to the model to estimate the masks in the rest of the frames in the video sequence. All models are trained during a fixed amount of 40 epochs.

%The evaluation metrics used to quantify the obtained results are 
We adopt the official metrics for the MOTS Challenge~\cite{voigtlaender2019mots} to obtain quantitative results: sMOTSA, MOTSP, Recall and Precision.
In all cases, the higher the metric, the better.
%They have been computed 
%Our results are computed over the validation set of the KITTI-MOTS dataset~\cite{voigtlaender2019mots}. 
However, instead of averaging the metrics per pixel as in the public benchmark, we have averaged them by sequence.
Otherwise, the results over longer sequences would dominate over the rest.

Two different strategies have been considered when allocating memory in the GPUs for training: whether we considered a lower spatial resolution (256x448 pixel) and longer clips of 5 frames, or a higher spatial resolution (287x950) and shorter clips of 3 frames.
While the 287x950 definition matches the aspect ratio of the KITTI-MOTS dataset~\cite{voigtlaender2019mots}, the 256x448 one corresponds to the aspect ratio of the YouTube-VOS dataset~\cite{xu2018youtube}, for which RVOS was originally trained.
%%ADDED%%----------------------------------------------------------

The KITTI-MOTS competition addresses a zero-shot challenge while our work has been focused on addressing a one-shot challenge. RVOS has demonstrated better performance with one-shot learning, which has been the motivation for choosing this approach. For this reason, the obtained results will not be compared with other state of the art works. Our objective is to explore the impact of the curriculum learning strategies on the performance of this model.
%%----------------------------------------------------------------

%because we have observed a large variance in terms of sequence length in 

%Due to the unique characteristics of each sequence and the unbalance of the number of frames, a new methodology for evaluation is proposed. The overall results presented on the tables are computed by averaging of the sMOTSA of the sequences over the total number of sequences instead of computing it on the pixel level for all sequences equally.

\subsection{Schedule Sampling}

Our experiments on schedule sampling consider the step and linear schedules in addition to the teacher forcing, provided as a baseline to compare with.
%We have considered four schedule sampling strategies in our experiments: the forward and inverse linear schedule and the forward and inverse step schedule.
%In addition, we have also considered the teacher forcing set up as a baseline to compare with.
%Five models have been trained to study the Schedule Sampling strategy. Two schedule sampling strategies are implemented on a forward and a inverse approach. The first strategy is the one that the original RVOS already implemented: Step schedule sampling. During the first half of training, the model only will be trained using the ground-truth annotations as input for the following step. On the second half of the training, the model will use its outputs. The other schedule sampling strategy studied is the linear schedule sampling. In this case, the model chooses if using ground-truth annotations or its outputs according to a threshold. It starts training with the ground-truth annotations and the probability of making this decision decreases as time goes by until, at the end, the model only chooses its outputs. This strategy is implemented with a threshold with decreases linearly. By generating a random number and checking whether it is higher or lower than the threshold, the model chooses either the ground-truth annotations or its outputs, respectively.
The study extends to the non-conventional inverse variations for both the step and linear cases, inspired by the finding reported in \cite{how_not_to}.
%To explore further the performance obtained with schedule sampling strategies and motivated by \cite{how_not_to}, inverse strategies for both the step and linear approaches have been implemented. 
The inverse variations actually defy the curriculum learning paradigm, as they start the training with the prediction of the model as references, and progress into a set up that considers only ground truth labels at the end.
%These strategies are implemented in the same way as in the forward approach with the difference that the model will start training with its outputs and, in the end, it will train with the ground-truth annotations. Also, to provide a global view, a teacher forcing strategy has been analysed. 
    
\begin{figure}[t]
        \centering
        \begin{tabular}{@{}c@{}}
            \includegraphics[width=\linewidth]{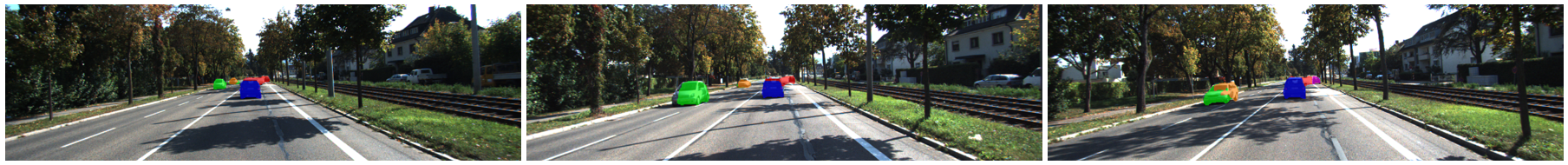} 
        \end{tabular}
        \begin{tabular}{@{}c@{}}
            \includegraphics[width=\linewidth]{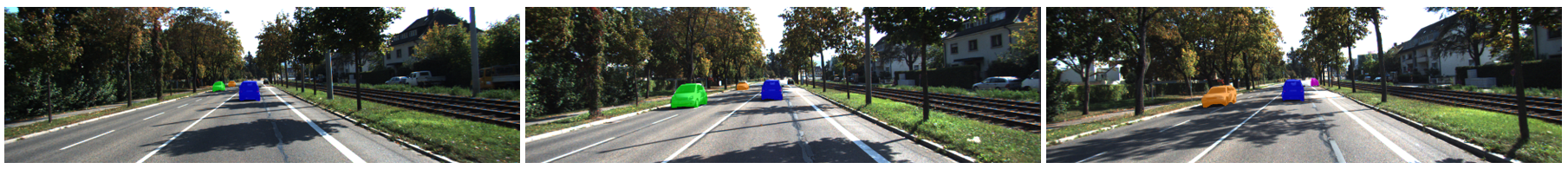} 
        \end{tabular}
    \caption{Qualitative results on three non-consecutive frames comparing the baseline model (row 1) and the model with the best performance: inverse step (row 2). Compared to the inverse step strategy, during all the sequence, on the baseline model, a wrong mask in red is observed next to the blue instance. Also, the orange mask is confused by the green mask.}
    \label{fig:qualitative_SS}
    \end{figure}

The results presented in Table \ref{tab:results_SS} indicate that actually the Forward Step curriculum adopted in the original RVOS baseline is the worst option, and that actually the best option is the inverse step approach.
%All models improve compared to the RVOS baseline, which adopted the Forward Step. 
%The best  which offers the best performance is the inverse step strategy with an image resolution of 287x950. 
Figure \ref{fig:qualitative_SS} shows a fragment of a sequence in which the inverse step outperforms the baseline model. 

    \begin{table}[t]
        \caption{Schedule sampling variations of one-shot VOS on KITTI-MOTS \textit{cars}. Best values are shown in \textbf{bold} and second best values in \color{blue}{blue}.}
        \centering
        \begin{tabular}{c|c|c|c|c|c|c|c|}
        \cline{2-8}
        & Image & Batch & Length &\multirow{2}{*}{sMOTSA}& \multirow{2}{*}{MOTSP} & \multirow{2}{*}{Recall} & \multirow{2}{*}{Precision}  \\ 
          &resolution & size & clip &  &  & &  \\ 
          
         \hline
         \multicolumn{1}{|c|}{\multirow{2}{*}{Teacher Forcing}} & 256x448 & 4 & 5 &-16.57 & 73.98 & 32.81 & 43.62 \\
         \multicolumn{1}{|l|}{}& 287x950 & 2 & 3 & \color{blue}{4.24} & 77.00 & 45.84 & \color{blue}{57.87} \\
         \hline
         
         \multicolumn{1}{|l|}{\multirow{2}{*}{Forward Step}} & 256x448 & 4 & 5 & -6.83 & 68.12 & 37.38 & 49.70 \\ 
         \multicolumn{1}{|l|}{}& 287x950 & 2 & 3 & -11.70 & 75.68 & 46.47 & 47.63 \\
         \hline
         
         \multicolumn{1}{|l|}{\multirow{2}{*}{Forward Linear}} & 256x448 & 4 & 5 & -2.29 & 72.97 & 41.00 & 53.64 \\ 
         \multicolumn{1}{|l|}{}& 287x950 & 2 & 3 & -5.58 & 76.76 & 46.72 & 51.53 \\ 
         \hline
         
         \multicolumn{1}{|l|}{\multirow{2}{*}{Inverse Step}}& 256x448 & 4 & 5  & -1.57 & 73.17 & 42.79 & 55.00 \\ 
         \multicolumn{1}{|l|}{}& 287x950 & 2 & 3 & \textbf{8.90} & \textbf{77.90} & 42.86 & \textbf{60.33}  \\
         \hline

         \multicolumn{1}{|l|}{\multirow{2}{*}{Inverse Linear}} & 256x448 & 4 & 5 & -4.77 & 73.35 & \textbf{48.60} & 53.06 \\ 
         \multicolumn{1}{|l|}{}& 287x950 & 2 & 3 & 2.48 & \color{blue}{77.87} & \color{blue}{47.12} & 57.07 \\ 
         \hline
        \end{tabular}
        
        \label{tab:results_SS}
        \end{table}

\subsection{Frame Skipping}

Two frame skipping schemes were explored. In the \textit{0 to 9} scheme, the number of skipped frames, which will be referred to as skipping step, is changed every 2 epochs. The total number of skipping steps is 10. The model starts training without skipping any frame and, gradually, increases the number of skipped frames by 1 until 9 consecutive frames are skipped. The second scheme, the \textit{1 to 5} one, halves the number of skipping steps from 10 to 5. In this case, the number of skipped frames is increased after 4 epochs, doubling the training time per skipping step.

These experiments are run with the RVOS baseline mode, which follows the Forward Step schedule sampling.    
%As said previously, the RVOS model is used. This model implements forward pulse schedule sampling. 
%For this reason, we also study the effect of implementing the frame skipping over the training epochs that use the ground truth (GT) or the predicted (Pred.) masks.
%when implementing frame skipping, the two training phases are differentiated. 
On the first training phase, when using the ground-truth (GT) annotations, frame skipping is always used. During the second training phase, when the model's predictions (Pred.) are used for training, we consider the two cases of skipping and non-skipping frames.
We consider this hybrid approach because the difficulty of having to deal with the noisy predictions of the model may be overwhelming for our model when adding on top the temporal sub-sampling.
%, as the difficulty for the model is already increasing due to the schedule sampling strategy, both with and without applying frame skipping has been studied. 
During the second phase, when frame skipping is applied, the skipping step begins from 0 and increases to 9 again.
    
The results in Table \ref{tab:results_FS} actually show that applying a frame skipping strategy during all training does not improve the performance of the model, maybe due to the difficulty of combining the two schemes.
Instead, when using frame skipping only during the first training phase, the performance improves considerably for either set of experiments. 
As the sequences of KITTI-MOTS present a slow motion, the model benefits from training with this scheme. 
Analysing the results for both configurations, it can be seen how the best results are obtained with a frame skipping scheme of increasing from 1 to 5 skipped frames. The model benefits more when seeing changes but with enough time to process them.
    
     \begin{table}[t]
        \caption{Frame skipping variations of one-shot VOS on KITTI-MOTS \textit{cars}. Best values are shown in \textbf{bold} and second best values in \color{blue}{blue}.}
        \centering
        \begin{tabular}{c|c|c|c|c|c|c|c|c|c|}
         \cline{2-10}
         &Image& Batch & Length & Skip  & Skip & \multirow{2}{*}{sMOTSA} & \multirow{2}{*}{MOTSP} & \multirow{2}{*}{Recall} & \multirow{2}{*}{Precision} \\
         &resolution& size & clip & @ GT & @ Pred. &  &  &  &  \\

         \hline
         \multicolumn{1}{|l|}{\multirow{2}{*}{No skip}} & 256x448 & 4 & 5  & No & No & -6,83 & 68,12 & 37,38 & 49,70 \\ 
         \multicolumn{1}{|l|}{}& 287x950 & 2 & 3 & No & No & -11,70 & 75,68 & 46,47 & 47,63 \\
         \hline
         \multicolumn{1}{|l|}{\multirow{4}{*}{0 to 9}} & 256x448 & 4 & 5  & Yes & Yes & -39,39 & 58,30 & 1,57 & 3,33 \\ 
         \multicolumn{1}{|l|}{}& 287x950 & 2 & 3  & Yes & Yes & -17,66 & 74,99 & 46,70 & 50,00 \\
         \cline{2-10}
         \multicolumn{1}{|l|}{}& 256x448 & 4 & 5  & Yes & No & \color{blue}{-0,87} & 74,73 & \color{blue}{49,43} & \textbf{55,49} \\ 
         \multicolumn{1}{|l|}{}& 287x950 & 2 & 3 & Yes & No & -8,18 & \color{blue}{76,92} & 44,67 & 48,21 \\ 
         
         \hline
         \multicolumn{1}{|l|}{\multirow{4}{*}{1 to 5}} & 256x448 & 4 & 5  & Yes & Yes & -43,44 & 70,43 & 27,16 & 32,06 \\
         \multicolumn{1}{|l|}{}& 287x950 & 2 & 3 & Yes & Yes & -22,87 & 75,20 & 41,77 & 45,99 \\ 
         \cline{2-10}
         \multicolumn{1}{|l|}{}& 256x448 & 4 & 5 & Yes & No & \textbf{0,51} & \textbf{79,10} & 39,26 & 53,57 \\ 
         \multicolumn{1}{|l|}{}& 287x950 & 2 & 3 & Yes & No & -7,05 & 75,86 & \textbf{53,00} & \color{blue}{54,49} \\
         \hline
         
        \end{tabular}
        
        \label{tab:results_FS}
        \end{table}
\begin{comment}    
    \begin{figure}[t]
        \centering
        \begin{tabular}{@{}c@{}}
            \includegraphics[width=\linewidth]{FS_0007_Baseline_qualitative.PNG}
        \end{tabular}
        \begin{tabular}{@{}c@{}}
            \includegraphics[width=\linewidth]{FS_0007_1to5_qualitative.PNG} 
        \end{tabular}
    \caption{Qualitative results on three non-consecutive frames depicting a turning scene comparing the baseline model (row 1) and the model with the best performance using frame skipping: skipping from 1 to 5 (row 2). The baseline model does not adapt well to the changes in position of the red segmented car, spreading the mask.}
    \label{fig:qualitative_FS}
    \end{figure}
\end{comment}
    
\subsection{Combination of Techniques}

The previous experiments were performed as isolated experiments to fully understand the impact of each technique over the baseline model, the Forward Step. After obtaining these results, one extra experiment has been studied with the best configurations of each technique. The combination of inverse step schedule sampling and frame skipping gives an overall sMOTSA of 16.05, outperforming the results of 8.9 and -7.05 given by the inverse step schedule sampling and frame skipping from 1 to 5 respectively. This experiment has been tested with the larger image resolution, as the performance on the inverse step with this configuration obtained the highest value among all the other experiments.

\section{Conclusions}

This work has shown how the curriculum learning greatly affects the performance of a deep neural network trained for the task of one-shot video object segmentation.
The two techniques explored, schedule sampling and frame skipping, have brought significant gains to the RVOS model.
These results encourage further research for a complete understanding and characterization of the techniques, especially in the surprising findings that an inverse step set up may result in better results.
However, the low values of the quantitative results also invite to explore these curriculum learning with better performing architectures that may produce more stable and confident results. Future work includes exploring these strategies in other datasets as well as further research on the combination of the strategies with the best results. %The results for the inverse schedule sampling question whether an inverse strategy with the frame skipping curriculum would improve the results so further investigations on this area are also left for future work.

%\clearpage

\section*{Acknowledgements}

This work was partially supported by the Spanish Ministry of Economy and Competitivity and the European Regional Development Fund (ERDF) under contract TEC2016-75976-R and RTI2018-095232-B-C22. We gratefully acknowledge the support of NVIDIA Corporation with the donation of GPUs.% used in this work.

%In this work, we have explo some useful strategies for the video object segmentation task with recurrent neural networks. It has been seen how ideas such as schedule sampling are worth exploring and interesting results have been presented when using inverse schemes such as inverse step schedule sampling. It has also been demonstrated that frame skipping strategies can take advantage of datasets with slow-motion scenes and improve the model's performance. 

\par\vfill\par

% ---- Bibliography ----
%
% BibTeX users should specify bibliography style 'splncs04'.
% References will then be sorted and formatted in the correct style.
%
\bibliographystyle{splncs04}
\bibliography{egbib}
\end{document}